\documentclass[twocolumn,10pt]{asme2e}

\confshortname{DSCC 2020}
\conffullname{the ASME 2020 \\ Dynamic Systems and Control Conference\\
Paper accepted for publication in DSCC2020}
\confdate{October 4-7}
\confyear{2020}
\confcity{Pittsburgh, Pennsylvania}
\confcountry{USA}

\usepackage{graphicx}
\usepackage{amsmath}
\usepackage{graphicx}
\usepackage{float}
\usepackage{pifont}
\usepackage{pgf}
\usepackage{tikz}
\usepackage{commath}
\usepackage{subfig}
\usepackage{epstopdf}
\usepackage[]{algorithm2e}
\usepackage{algorithmic}
\usepackage{ifthen}
\usepackage{subfig}
\usepackage{bbm}
\usepackage{dsfont}
\usepackage{amssymb}
\usepackage{txfonts}
\usepackage{mathbbol}
\usepackage{multirow}
\usepackage{lipsum}

\usepackage{makecell}

\papernum{DSCC2020-3319}

\title{Developmental Reinforcement Learning of Control Policy of a Quadcopter UAV with Thrust Vectoring Rotors}

\author{Aditya M. Deshpande, Rumit Kumar
    \affiliation{
	Cooperative Distributed Systems Lab\\
	University of Cincinnati, Cincinnati, Ohio 45221, USA\\
    Email: $\lbrace$ deshpaad, kumarrt $\rbrace$ @mail.uc.edu
    }}	
    
\author{Ali A. Minai and Manish Kumar
    \affiliation{
    College of Engineering and Applied Sciences\\
	University of Cincinnati, Cincinnati, Ohio 45221, USA\\
    Email: $\lbrace$ ali.minai, manish.kumar $\rbrace$ @uc.edu
    }
}

\begin{document}
\maketitle
\pagenumbering{roman}  
\setcounter{page}{1}
\begin{abstract}
\noindent In this paper, we present a novel developmental reinforcement learning-based controller for a quadcopter with thrust vectoring capabilities. This multirotor UAV design has tilt-enabled rotors. It utilizes the rotor force magnitude and direction to achieve the desired state during flight. The control policy of this robot is learned using the policy transfer from the learned controller of the quadcopter (comparatively simple UAV design without thrust vectoring). This approach allows learning a control policy for systems with multiple inputs and multiple outputs. The performance of the learned policy is evaluated by physics-based simulations for the tasks of hovering and way-point navigation. The flight simulations utilize a flight controller based on reinforcement learning without any additional PID components. The results show faster learning with the presented approach as opposed to learning the control policy from scratch for this new UAV design created by modifications in a conventional quadcopter, i.e., the addition of more degrees of freedom (4-actuators in conventional quadcopter to 8-actuators in tilt-rotor quadcopter). We demonstrate the robustness of our learned policy by showing the recovery of the tilt-rotor platform in the simulation from various non-static initial conditions in order to reach a desired state. The developmental policy for the tilt-rotor UAV also showed superior fault tolerance when compared with the policy learned from the scratch.
The results show the ability of the presented approach to bootstrap the learned behavior from a simpler system (lower-dimensional action-space) to a more complex robot (comparatively higher-dimensional action-space) and reach better performance faster.
\end{abstract}

\vspace{-18pt}

\section{Introduction}
\noindent With the increase in applications of unmanned aerial vehicles (UAVs) in the civilian domain, there is also a surge in the development of novel UAV designs based on evolving operational requirements. In this space of aerial vehicles, the multirotor platforms are the most popular and have been useful in applications including but not limited to terrain photogrammetry \cite{gonccalves2015uav}, agriculture \cite{das2015devices}, power line inspection \cite{luque2014power}, package delivery \cite{stolaroff2018energy}. Various applications and challenges have resulted in the development of several advanced UAV designs. Conventional quadcopters, tethered UAV configurations, variable blade pitch quadcopters, morphological aerial platforms, hexacopters and tilt-rotor quadcopters are some of the popular names.
\begin{figure}[t]
	\centering
	\includegraphics[scale=0.58]{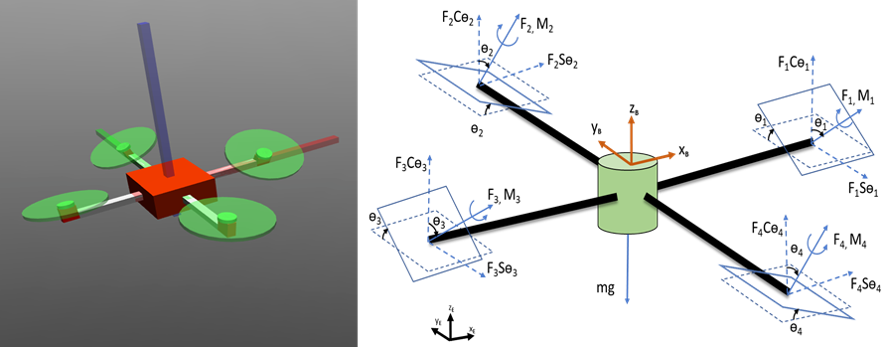}
	\vspace{-3pt}
    \caption{Tilt-rotor quadcopter model. The $RGB$-colored body-axes on mujoco model (left) correspond to $XYZ$-axes respectively on the free-body diagram (right).}
    \label{fig1-tilt-rotor-mujoco}
    \vspace{-19pt}
\end{figure}
\newline
\indent Quadcopters are among the most studied and adopted platforms \cite{hassanalian2017classifications} in UAV robotics. The quadcopter platform is an under-actuated system with four controlling rotors. It is controlled by modulating the angular speeds of the rotors based on a full state feedback control law. Most of the novel UAV designs are derivations or extensions of the conventional platform. A tethered quadcopter is a platform designed for long-endurance flights \cite{nicotra2017nonlinear}. It uses a taut cable during the flight to connect with the continuous power supply on the ground. Similarly, the variable pitch quadcopter is another interesting design that can perform aggressive flight maneuvers and even achieve inverted flight modes \cite{cutler2015analysis, sheng2016control, gupta2016flight}. Tail-sitter UAV is another design that can fly like a fixed-wing aircraft while take-off and landing are done similar to multi-rotor UAV \cite{swarnkar2018biplane, ritz2018global}. It uses wing aerodynamics for lift generation during cruise flight. A morphing quadcopter design that can change shapes during the flight is presented by Falanga et al. \cite{falanga2018foldable}. This design includes four additional servo actuators used for morphing UAV shape along with four motors used to generate thrust during the flight. The adaptive morphing capability of this UAV design increased the versatility of this platform. The experiments presented in this work showed the capability of this platform in tasks such as navigation through narrow spaces, object grasping and transportation. A novel passive morphing design and control of quadcopter was presented in \cite{bucki2019design}, this UAV was designed with sprung hinges on its arms fold downwards when low thrust commands were applied. The work in \cite{zhao2018design} presented yet another design derived from quadcopter with additional degrees of freedoms which enabled aerial manipulation. Another variation of quadcopter design and its control is presented in \cite{kumar2020flight} where $dof$s are introduced in the quadcopter arms.
The platform designed and studied in this work was called \textit{DRAGON} which can transform its shape in-flight using multiple links on its body. Tilt-rotor is an over actuated form of the quadcopter design \cite{kumar2018reconfigurable}. These systems have eight actuators inputs and can provide control over each independent degree of freedom. In this paper, we have focused on the developmental reinforcement learning of control policy for the tilt-rotor UAV. In previous works, the controller synthesis of the tilt-rotor UAV has been explored by many researchers. The mathematical dynamic modeling of the tilt-rotor UAV are discussed in \cite{ryll2015novel}.
The PD based control methods and non-linear sliding mode controller design for the tilt-rotor UAV were studied in \cite{kumar2020quaternion, sridhar2018fault, kumar2017tilting}. This system has also been described as capable of handling single motor failure during flight \cite{sridhar2018fault}. Invernizzi in \cite{invernizzi2018trajectory} used geometric control theory for the controller development of the tilt-rotor UAV. 

\begin{figure*}[t]
	\centering
	\includegraphics[scale=0.8]{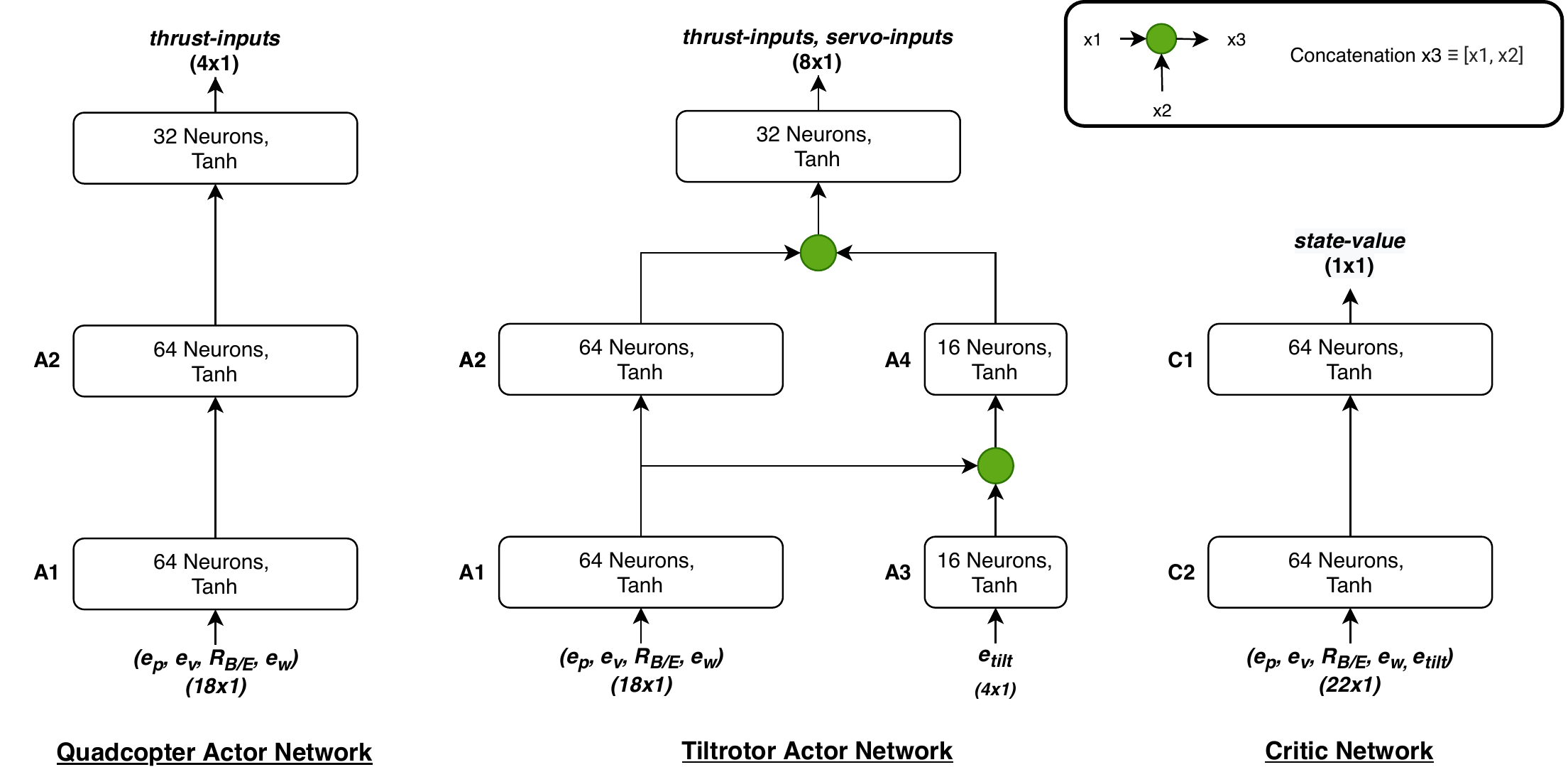}  
	\vspace{-5pt}
    \caption{Neural Network Architectures: $Quadcopter$ $Actor$ $Network$ (left) is the control policy used for Quadcopter; $Tiltrotor$ $Actor$ $Network$ (center) is the policy architecture used for Tilt-rotor quadcopter; $Critic$ $Network$ (right) acts as the state-value function for an observed state of tilt-rotor.}
    \label{policy-arch}
    \vspace{-15pt}
\end{figure*}

Model predictive control (MPC) is another effective method for controlling the aerial robotic systems \cite{mueller2013model, zhang2016learning}, but the computational cost of MPC-based controllers can be high as it requires estimating the state of the system. For a robot with many degrees of freedom, state estimation is a non-trivial problem. With deep reinforcement learning, we can in principle avoid state estimation problem and learn a control policy which directly maps sensor data to control actions \cite{lillicrap2015continuous}.

With the increase in the number of actuators or design modifications of the quadcopter, the control law formulation for the system becomes challenging and sometimes even intractable. Instead of designing a control law, it may be feasible to learn the control policy for these advanced systems. In recent years, deep reinforcement learning (RL) based approaches have emerged as a promising method for controlling systems with high-dimensional action-spaces \cite{SergeyFinnEnd2016, ishige2019exploring, recht2019tour}. Using the learning-based approach, the system can explore the state-space effectively and learn the control policies which may be nearly impossible to design using traditional linear and non-linear control methods due to the increased complexity of the dynamics \cite{zeng2019tossingbot}. With this motivation, we investigate learning control policies for the over actuated tilt-rotor UAV. We consider a developmental learning approach to learn a feasible control policy with increasing action-space of the UAV due to design modifications. Instead of learning a control policy from scratch, we allow the robot to learn in stages starting from a policy for the basic system and then moving on to a more advanced system. In this work, we consider the conventional quadcopter with four motor inputs as the basic system while the tilt-rotor UAV with eight actuator inputs is considered as the advanced system.

The RL algorithms which have achieved state-of-the-art results on various robots rely on deep neural networks because of their good approximation capabilities. With advancements in computing and auto differentiation frameworks, there is a surge in the adoption of neural networks in the robotics community. They have even found applications in learning control policies for the multirotor UAVs. Hwangbo et al. \cite{hwangbo2017control} present a quadcopter policy learning approach in the actor-critic framework using deterministic policy optimization. The RL algorithm was developed based on information-theoretic natural gradient descent. The algorithm presented in this work is conservative and uses a PID controller along with a learned policy to stabilize the learning process. Koch et al. presented an RL-based approach to learn the quadcopter attitude controller in \cite{Koch2019UAVatt}. Another excellent work on the sim-to-real transfer of learned UAV policies using deep RL was presented by Molchanov et al. in \cite{sim2multireal2019}. Here, the authors showed a successful transfer of quadcopter control policies learned in simulation to the real-world platforms. They considered an $x$-configuration quadcopters with varying physical properties (such as mass and dimensions of the system) to analyze the robustness of the learned policies.
\newline
\indent Developmental robotics is one of the branches in artificial intelligence concerned with robot adaptation and learning through qualitative growth of behavior and increasing levels of competence. It deals with the development of embodied agents using algorithms that allow life-long and open-ended learning \cite{morse2010epigenetic}.
So far, there has been very limited work on developmental learning in multirotor UAVs. In this work, we apply a developmental learning approach in UAVs with advancements in their design. This work is inspired by the developmental reinforcement learning approach presented in \cite{zimmer2018developmental}. Our approach begins with the search for a feasible control policy in a smaller search space (conventional quadcopter) with the initial objective to reach the desired state. This approach then bootstraps the knowledge gained by the conventional quadcopter gained after its training.
We test this approach using an on-policy, model-free RL algorithm of proximal policy optimization (PPO) \cite{ppo2017schulman}.

\vspace{-16pt}

\subsection{Contributions}
\noindent This paper presents a novel approach based on developmental reinforcement learning to learn a control policy for the tilt-rotor quadcopter. The approach presented in this work is a type of curriculum learning which seeks to speed up the learning of a robot with a large number of degrees of freedom (\emph{dof}) by first learning to control a simpler robot design with comparatively fewer \emph{dof} \cite{narvekarCurriculum2019}.
The contribution of this work is two-fold. First, we demonstrate the deep reinforcement learning based control of a tilt-rotor UAV platform. Although RL algorithms for learning control policies for quadcopters have been explored before, there has been limited work in exploring these approaches for controlling more advanced UAV designs. We use a neural network for policy approximation in this work. Second, we show that learning in stages as presented in our approach can lead to qualitatively better policies in fewer iterations for high-dof systems. We evaluate our approach using the tilt-rotor quadcopter platform shown in fig. \ref{fig1-tilt-rotor-mujoco}. We developed our training environment using Mujoco Physics \cite{todorov2012mujoco} and OpenAI Gym \cite{gym2016openai}. The performance of the learned policy is compared with a PID-controller.

\vspace{-15pt}

\section{Dynamic Model} \label{sec2}
\noindent In this section, we briefly discuss the dynamics of the tilt-rotor quadcopter. A quadcopter of `$Plus$'-configuration is used in the work. The dynamic model of the system is based on the work in \cite{kumar2018reconfigurable}. We have reproduced the equations of motion for the sake of completeness. The free body diagram ofthe  tilt-rotor quadcopter is shown in fig. \ref{fig1-tilt-rotor-mujoco}.  The reader should note that the sine and cosine angle terms are represented as $s\angle$ and $c\angle$ respectively. The translational motion of UAV in the world frame is represented by equation \eqref{translation-motion}. The rotation matrix $R_{B/E}$ rotates the body frame ($B$) parameters to the world frame ($E$). The body-fixed frame $B$ is attached to the center of gravity of the vehicle. This rotation matrix defines a $Z-Y-X$ Euler angle transformation by $\psi$-$\theta$-$\phi$ angles representing yaw-pitch-roll respectively. Rotor tilt angles are defined by $\theta_i$, $\forall i \in {1,2,3,4}$ and the thrust forces produced by each rotor are denoted by $F_i$, $\forall i \in {1,2,3,4}$. The inputs to this system are the tilt-angle rates $\dot \theta_i$ and motor thrusts $F_i$.
The mass of the system is represented by $m$, $g$ represents acceleration due to gravity, and $\ddot{x}$, $\ddot{y}$, $\ddot{z}$ represent the acceleration of the system in world frame $E$. Equation \eqref{rotation-motion} represents the rotational motion of the tilt-rotor system. $I$ is a diagonal matrix containing terms of moments of inertia about $x_B, y_B, z_B$-axes. The quadcopter arm length is given by $l$.

\vspace{-15pt}

\begin{eqnarray}
\begin{bmatrix} 
\ddot{x}\\\\
\ddot{y}\\ \\
\ddot{z} 
\end{bmatrix}  
&=&\frac{R_{B/E}}{m}
\begin{bmatrix} 
F_2s\theta_2 + F_4s\theta_4\\\\
-F_1s\theta_1 - F_3s\theta_3\\\\
F_1c\theta_1 + F_2c\theta_2 + F_3c\theta_3 + F_4c\theta_4
\end{bmatrix} 
-
\begin{bmatrix} 
 0\\\\
 0\\\\
 g
\end{bmatrix}
      \label{translation-motion}\\
\nonumber\\
I
\begin{bmatrix} 
\dot{p}\\
\\\\
\dot{q}\\
\\\\
\dot{r} 
\end{bmatrix}  
&=&
\begin{bmatrix} 
l(F_2c\theta_2 - F_4c\theta_4) + M_2s\theta_2 - M_4s\theta_4\\\\
l(F_3c\theta_3 - F_1c\theta_1)- M_3s\theta_3 + M_1s\theta_1\\\\
l(-F_1s\theta_1 - F_2s\theta_2+ F_3s\theta_3 + F_4s\theta_4)\\
- M_1c\theta_1 + M_2c\theta_2 + M_3c\theta_3 - M_4c\theta_4
\end{bmatrix}
-
\begin{bmatrix} 
p\\\\
\\
q\\\\
\\
r
\end{bmatrix}
\times I
\begin{bmatrix} 
p\\\\
\\
q\\\\
\\
r
\end{bmatrix} 
      \label{rotation-motion}
\end{eqnarray}

\vspace{-7pt}

\noindent The roll, pitch and yaw rates are given by $p$, $q$, $r$  in the quadcopter body frame, respectively. The torque produced as a result of angular motion of propellers is given by $M_i$, $\forall{i}\in\lbrace 1,2,3,4\rbrace$.  The Euler angle rates are obtained from body angular rates as discussed in \cite{kumar2017tilting} and the Euler angles can be computed by numerical integration of Euler angle rates. The thrust force and the torque produced by each motor is directly proportional to the square of angular speed of the propeller with the constants of proportionality given by $k_f$ and $k_m$, respectively \cite{grasp}.

\vspace{-16pt}

\section{Policy Training}
\noindent This section provides detailed information on the training of the neural network policy using the developmental approach.

\vspace{-15pt}

\subsection{Background} \label{background}
\noindent The reinforcement learning problem is modeled as a Markov decision process (MDP) which is described by the tuple $\langle S, A, P, R, \gamma \rangle$ where $S$ is the state space, $A$ is the action space, $P(s_{t+1}|s_{t}, a_{t})$ is the probability distribution over state transitions, $R(s_{t}, a_{t}, s_{t+1})$ is the reward function, and $0 \le \gamma < 1$ is the discount factor \cite{sutton2018reinforcement}. In this work, we consider a continuous state space (or observation space) and a continuous action space for the robot. The RL objective is to learn a control policy $\pi(a_{t}\mid s_{t})$ that can maximize the expected cumulative reward. The initial state $s_0$ of the RL agent is sampled from a fixed distribution $p(s_0)$. At each time step $t$, the agent takes action $a_{t}\sim \pi(\cdot \mid s_{t})$ where $a_{t} \in A$ to go from state $s_{t}$ to state $s_{t+1} \sim P(\cdot|s_{t}, a_{t})$. The agent gets a reward $r_{t} = R(s_{t}, a_{t}, s_{t+1})$ after each action. Equation \eqref{policy-obj} defines the objective where $\tau = (s_0, a_0, s_1, a_1, ...)$ is the trajectory sampled using policy $\pi$.

\vspace{-27pt}

\begin{equation}
\pi^{*} = \arg\max_{\pi} \mathbb{E}_{\tau \sim \pi} \begin{bmatrix} R(\tau) \end{bmatrix} = \arg\max_{\pi} \mathbb{E}_{\tau \sim \pi} \begin{bmatrix} \sum_{t=0}^{\infty} \gamma ^ t r_t \mid \pi \end{bmatrix} 
\label{policy-obj}
\end{equation}

\vspace{-30pt}

\subsection{Proximal Policy Optimization (PPO)} \label{section-ppo}
\noindent We use the PPO algorithm to train our policy to optimize the objective described in equation \eqref{policy-obj}. PPO is an on-policy algorithm and it trains the stochastic policy. Thus, the RL agent is able to do the exploration by sampling from its latest version of the stochastic policy. The output of the policy is the mean of a multi-variate Gaussian distribution with dimensions as the number of actuators on the system. We sampled the actions from this distribution with a constant diagonal covariance matrix with elements of equal magnitude $\sigma^2$ (refer table \ref{hyperparameter_table}). We made this choice for sampling the actions given the current observation of the agent to encourage its exploration throughout the training process and to avoid the policy getting trapped in local optima. Our implementation of PPO is based on the one available in \cite{pytorchrl}.

As typically done in the robotics literature, we assume that accurate observations of the system are available during training. Reasonably accurate state estimation of the robot is possible using localization achieved by the fusion of various sensor readings on the robot which may include the following: inertial measurement units, depth cameras, RGB-cameras, LIDAR, GPS, or an external motion capture system \cite{cully2015robots, sim2multireal2019}.
Noise was inserted in the initial state of the environment corresponding to the distribution $p(s_{0})$ described in section \ref{background}. This noise enables the RL agent to learn the task independent of the initial state of the robot. We followed the sampling strategy from \cite{sim2multireal2019} to initialize the environment for each episode during the training. As mentioned in section \ref{section-reward}, the desired location in the world coordinate frame was kept as $(0, 0, 5m)$  throughout the training. The drone was trained to hover at the desired location. The initial position was sampled uniformly from the cube spanning $2m$ around this location. The magnitude of the initial velocity (in $m/s$) was sampled from a uniform distribution in the range of $[0, 1]$ . The magnitude of angular velocity (in $rad/s$) was also sampled from a uniform distribution in range of $[0, 1]$ . We noticed that sampling the orientation of the UAV uniformly in $SO(3)$ made the training difficult and could take longer in converging to a stable policy. This is because the system may not be able to recover from certain initial states such as completely upside down initial orientation of the drone. Thus, to encourage exploration at the beginning of training, the samples were drawn from $SO(3)$ for the first $500$ episodes and then the range of roll, pitch and yaw angles was shrunk to $[-\frac{\pi}{3}, \frac{\pi}{3}]$ for rest of the training period. Each episode in training was terminated if its length exceeded the maximum number of time-steps set in the simulation. Additionally, the episode was terminated if the robot went beyond the bounds defined in the simulation. These bounds were defined as the cube of $3m$ around its target position.

\vspace{-15pt}

\subsection{Policy Architecture}\label{section-policy}
\noindent Figure \ref{policy-arch} illustrates the architecture of the neural network used in this work. Policy training is done using the actor-critic framework \cite{ppo2017schulman}. This is not an optimum architecture. For the presented work, we did not try other network architectures and did not consider the effect of varying the number of neurons or their activation functions. 
\newline
\indent The actor-network is the policy used to control the UAV. The critic-network is used to evaluate the value of the state of the UAV. We follow a simple transfer learning heuristic for accommodating the additional tilt-servos control-inputs of the tilt-rotor UAV after the performance of the base system saturates for the defined task. The training is done in two stages. In the first stage, only the conventional quadcopter with four motor inputs is trained using the $Quadcopter$ $Actor$ $Network$ [left-most in fig.\ref{policy-arch}]. When the performance for the given task for the base system (quadcopter) saturates over the training, the additional tilt-servo inputs are released in the system. Thus, the weights learned in layers $\mathbf{A1}$ and $\mathbf{A2}$ in quadcopter actor are transferred to corresponding layers in $Tiltrotor$ $Actor$ $Network$ [center in fig.\ref{policy-arch}] and are kept frozen during training.
Each network uses the hyperbolic tangent ($Tanh$) activation function. The use of $Tanh$ activation in the output neurons is convenient to keep the network outputs bounded within the range $[-1, 1]$. 
The critic network architecture [right-most in fig.\ref{policy-arch}] is similar to that of the quadcopter actor where layers $\mathbf{C1}$ and $\mathbf{C2}$ have the same structure as that of layers $\mathbf{A1}$ and $\mathbf{A2}$. The output of the critic network is a scalar. This scalar determines the value of the current state of the robot. It should be noted that in case of the quadcopter, the input vector of the critic network is 18-dimensional (not shown in fig.\ref{policy-arch}) while in the case of the tilt-rotor quadcopter, the input vector is a 22-dimensional observation vector.

\vspace{-13pt}

\subsection{Observation Space and Action Space}
\noindent The control policy learned using RL maps the current state of the system to the actuator commands. The observations used in the quadcopter include $(e_{p}, e_{v}, R_{B/E}, e_{\omega})$. 
The observations passed to the neural networks consist of the tuple of errors in the desired state of the system given by $(e_{p}, e_{v}, R_{B/E}, e_{\omega})$ where $e_{p} \in \mathbb{R}^{3}$ is the error in desired position, $e_v \in \mathbb{R}^{3}$ is the error in desired velocity, $e_{\omega} \in \mathbb{R}^{3}$ is the error in body rates, and $R_{B/E}$ is the flattened $3 \times 3$ rotational matrix ($R_{B/E}$ converted to a $9$-dimensional vector by flattening). Thus, the state space of the system is of 18 dimensions. 
For tilt-rotor UAV, in addition to the 18-dimensional observations as in case of the quadcopter, there are four more observations given by $e_{_{tilt}}\in\mathbb{R}^{4}$ that correspond to errors in rotor tilt angles. Thus, the observation space of the tilt-rotor system is 22-dimensional.

The dimension of the action space of the system under consideration is equal to the number of active actuators. For the quadcopter, the action space is $a \in \mathbb{R}^4$ corresponding to its four motors while in case of tilt-rotor the action space is $a \in \mathbb{R}^8$ since it also has four additional servos for tilting. The neural network policies have output dimensions equal to the number of actuators in the respective systems. We scale the policy outputs from [-1, 1] to the corresponding actuator limits. The output values are centered at the hovering condition of the quadcopter given by the equation \eqref{hovering-condition}.

\vspace{-25pt}

\begin{equation}
F_h = \frac{mg}{4}
\label{hovering-condition}
\end{equation}

\vspace{-10pt}

\noindent where $F_h$ is the required thrust force by each rotor for hovering state of the quadcopter \cite{grasp}. We use a linear mapping to go from neural policy output to actuator thrust given by \eqref{prop-out-scaling}.

\vspace{-20pt}

\begin{equation}
F_{i} = F_h + \frac{ a_{j} (F_{max} - F_{min})}{2}
\label{prop-out-scaling}
\end{equation}

\vspace{-15pt}

\noindent where $F_{min}$ and $F_{max}$ are minimum and maximum thrust forces possible for the actuators respectively, $i \in \{1,2,3,4\}$ represents the thrust actuator index, and $j \in \{1,2,3,4\}$ represents the index of output neuron of the policy. Similarly, tilt-servos inputs are scaled as shown in \eqref{tilt-out-scaling}.

\vspace{-20pt}

\begin{equation}
\dot \theta_{i} = \frac{ a_{j} (\dot\theta_{max} - \dot\theta_{min}) }{2}
\label{tilt-out-scaling}
\end{equation}

\vspace{-15pt}

\noindent where $\dot \theta_{min}$ and $\dot \theta_{max}$ are minimum and maximum tilt rates of the tilt-servos respectively, $i \in \{1,2,3,4\}$ represents the tilt-actuator index, and $j \in \{5, 6, 7, 8\}$ represents the corresponding index of action output from the tilt-rotor policy.

\vspace{-12pt}

\subsection{Reward Function}
\label{section-reward}
\noindent The robot is trained to reach the desired waypoint from its current position. This waypoint is defined in the world coordinate frame. The desired values of attitude and body rates are set to zero as the UAV reaches the goal location. The desired linear velocity of the system is also zero as it reaches the goal location. Based on these requirements we define the reward earned by the conventional quadcopter during each time step in equation \eqref{reward-func}.

\vspace{-15pt}

\begin{equation}
r_{t}= \beta - \alpha_{a}\|a\|_2 - \sum_{_{k\in\{p,v,\omega\}}}\alpha_{k} \|e_{k}\|_2 - \sum_{_{j\in\{\phi,\theta\}}}\alpha_{j}\|e_{j}\|_2
\label{reward-func}
\end{equation}

\vspace{-15pt}

\noindent where $\beta \ge 0$ is a bonus value earned by the agent for staying alive in the simulation and $\alpha_{\{\cdot\}} \ge 0$ represent weights for various terms in the reward function. The second term in \eqref{reward-func} represents the penalty for the actions of the robot. The third and fourth terms are the summation of penalties imposed due to errors in the state of the robot. The third term consists of the error in position ($e_p$), the error in velocity ($e_v$) and the error in body rates ($e_\omega$) of the system. Similarly, the last term consists of the error in roll ($e_\phi$) and the error in pitch ($e_\theta$) angles, respectively. The system was not penalized for error in yaw angle to prioritize the learning of policy to reach the desired waypoint irrespective of the UAV heading.
\newline
\indent For tilt-rotor UAV, the desired tilt angles of servos are set to zero. To account for tilt-angle error in the reward function we add an additional term in equation \eqref{reward-func} which results in equation \eqref{reward-func2} for tilt-rotor platform.

\vspace{-20pt}

\begin{equation}
r_{t}= \beta - \alpha_{a}\|a\|_2 - \sum_{_{k\in\{p,v,\omega\}}}\alpha_{k} \|e_{k}\|_2 - \sum_{_{j\in\{\phi,\theta\}}}\alpha_{j}\|e_{j}\|_2 - \alpha_{_{tilt}} \|e_{_{tilt}}\|_{2}
\label{reward-func2}
\end{equation}

\vspace{-15pt}

\noindent where $\alpha_{_{tilt}} \ge 0$ is a weight corresponding to error in servo tilts $e_{_{tilt}}$. All the remaining terms in this equation have the same meaning as explained for equation \eqref{reward-func}.

\vspace{-12pt}

\subsection{Policy Transfer} \label{section-policy-transfer}
\noindent In the presented approach we bootstrap the policy learned by the conventional quadcopter to further train the tilt-rotor UAV. 

First the quadcopter control policy is trained using the quadcopter actor shown in figure \eqref{policy-arch}. Once the quadcopter is trained, the parameters of layer $\mathbf{A1}$, $\mathbf{A2}$ are transferred to corresponding layers in tilt-rotor actor and kept frozen, i.e., no gradient is calculated for these weights while training the control policy for tilt-rotor quadcopter. The remaining parameters of the tilt-rotor quadcopter actor network were initialized using $Xavier$ initialization to values in the range $[-0.1, 0.1]$ \cite{glorot2010understanding}.

The training of the critic network was started from its weights obtained after the quadcopter training. The weights corresponding to connections between neurons in layers $\mathbf{C1}$ and $\mathbf{C2}$ and the weights corresponding to connections between neurons in layers $\mathbf{C2}$ and the output were transferred from the quadcopter critic. As the number of observations change for the tilt-rotor platform, the weights in the input layer were initialized using $Xavier$ initialization to values between range $[-0.1, 0.1]$. It should be noted that the critic was allowed to update its weights during the training of the tilt-rotor UAV.

\begin{figure}[]
	\centering
	\includegraphics[width=\linewidth]{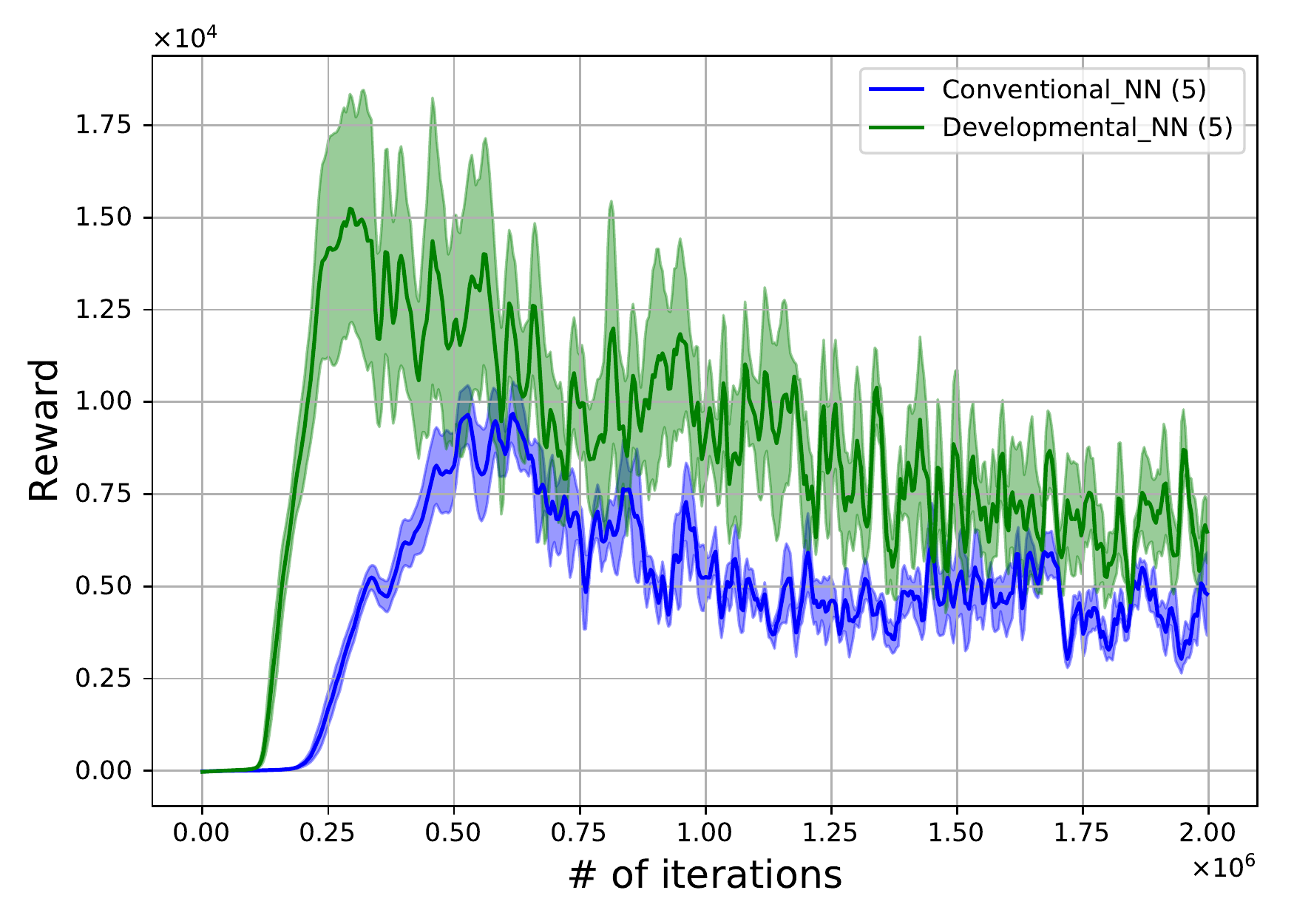}
	\vspace{-25pt}
    \caption{Comparison of reward plots for tilt-rotor quadcopter training using $Developmental-NN$ and $Conventional-NN$.}
    \label{reward_plot}
    \vspace{-5pt}
\end{figure}

\begin{table}[ht!]
    \centering
    \vspace{-10pt}
    \caption{UAV physics model parameters \label{modalparameters_table}}
    \vspace{-5pt}
    \begin{tabular}{||c|c||} 
         \hline
         Parameter & Value \\ [0.5ex] 
         \hline
         Timestep & 0.01 sec \\ [-0.5ex]
         Motor lag & 0.05 sec \\ [-0.5ex]
         Mass $m$ & 1.5 kg \\ [-0.5ex]
         Arm length $l$ & 0.13 m \\ [-0.5ex]
         Tilt angle range $[\theta_{min}, \theta_{max}]$ & $[-\frac{\pi}{3}, \frac{\pi}{3}]$ rad \\ [-0.5ex]
         Motor thrust range $[F_{min}, F_{max}]$ & $[0, 15.0]$ N \\ 
         \hline
    \end{tabular}
    \vspace{-15pt}
\end{table}

\vspace{-20pt}

\section{Simulation Setup}
\noindent To train the tilt-rotor UAV using RL we designed $multi-rotor gym$ environment using the Mujoco physics engine and OpenAI Gym. Two multi-rotor UAV designs, viz., quadcopter and tilt-rotor platforms, were simulated in this environment. The environment was modeled at sea-level atmospheric conditions. The tilt joints on the tilt-rotor quadcopter were modeled using velocity servos. Table \ref{modalparameters_table} provides the physical parameters of the robot used in this work. The values of coefficients in equations \eqref{reward-func} and \eqref{reward-func2} are: $\alpha_{p}=1.0$, $\alpha_{v}=0.05$, $\alpha_{\omega}=0.25$, $(\alpha_{\phi}, \alpha_{\theta})=(0.1, 0.1)$, $\beta=5.0$, $\alpha_{_{tilt}}=0.5$, $\alpha_{{a}}=0.25$. The simulations were performed on a machine with an Intel-i7 processor, 16GB RAM, and NVIDIA RTX 2070. The training of each environment took approximately $35$ minutes. The hyperparemeters used for training the RL policies are given in table \ref{hyperparameter_table}.

\vspace{-10pt}

\begin{table}[ht!]
    \centering
    \caption{Hyperparameters used to train the control policies \label{hyperparameter_table}}
    \vspace{-5pt}
    \begin{tabular}{||c|c||} 
         \hline
         Parameter & Value \\ [0.5ex] 
         \hline
         Training iterations & $2 \times 10^{6}$ \\ [-0.5ex]
         Learning rate & 5e-5 \\ [-0.5ex]
         Neural network optimizer & Adam \cite{kingma2014adam} \\ [-0.5ex]
         Adam parameters $(\beta_1, \beta_2)$ & $(0.9, 0.999)$ \\ [-0.5ex]
         Discount factor $\gamma$ & $0.95$ \\ [-0.5ex]
         Max. episode length & $1500$ \\ [-0.5ex]
         Minibatch size & $32$ \\ [-0.5ex]
         Clipping parameter for PPO & $0.2$ \\ [-0.5ex]
         GAE parameter $\lambda$ & $0.95$ \\ [-0.5ex]
         Epochs per PPO update & 10 \\ [-0.5ex]
         State-value loss coefficient & 0.5 \\ [-0.5ex]
         Variance of action distribution $\sigma^2$ & 1.0 \\ [-0.5ex]
         Learning rate schedule & linear\\
         \hline
    \end{tabular}
    \vspace{-10pt}
\end{table}

\vspace{-8pt}

\section{Results and Discussion}

\begin{figure*}[]
	\centering
	\vspace{0pt}
	\includegraphics[scale = 0.6]{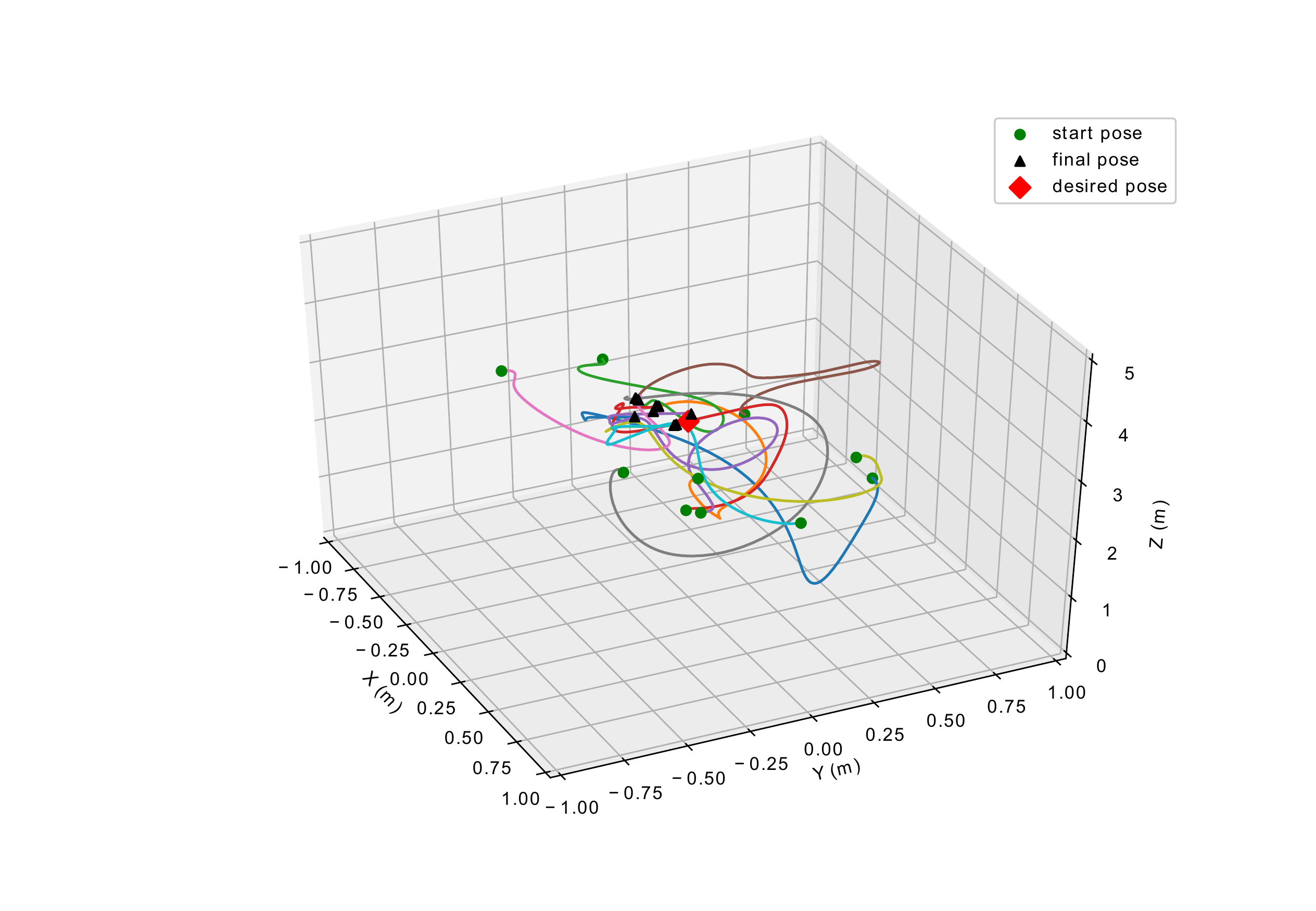}
	\vspace{-30pt}
    \caption{Illustration of tilt-rotor quadcopter trajectories from 10 randomly initialized location to final goal.}
    \label{trajectory-plot}
    \vspace{-15pt}
\end{figure*}

\begin{figure*}
\centering
\subfloat[Variation of UAV position \label{xyz-trajectory-plot}]{
  \includegraphics[width=0.48\textwidth, height=0.35\textwidth]{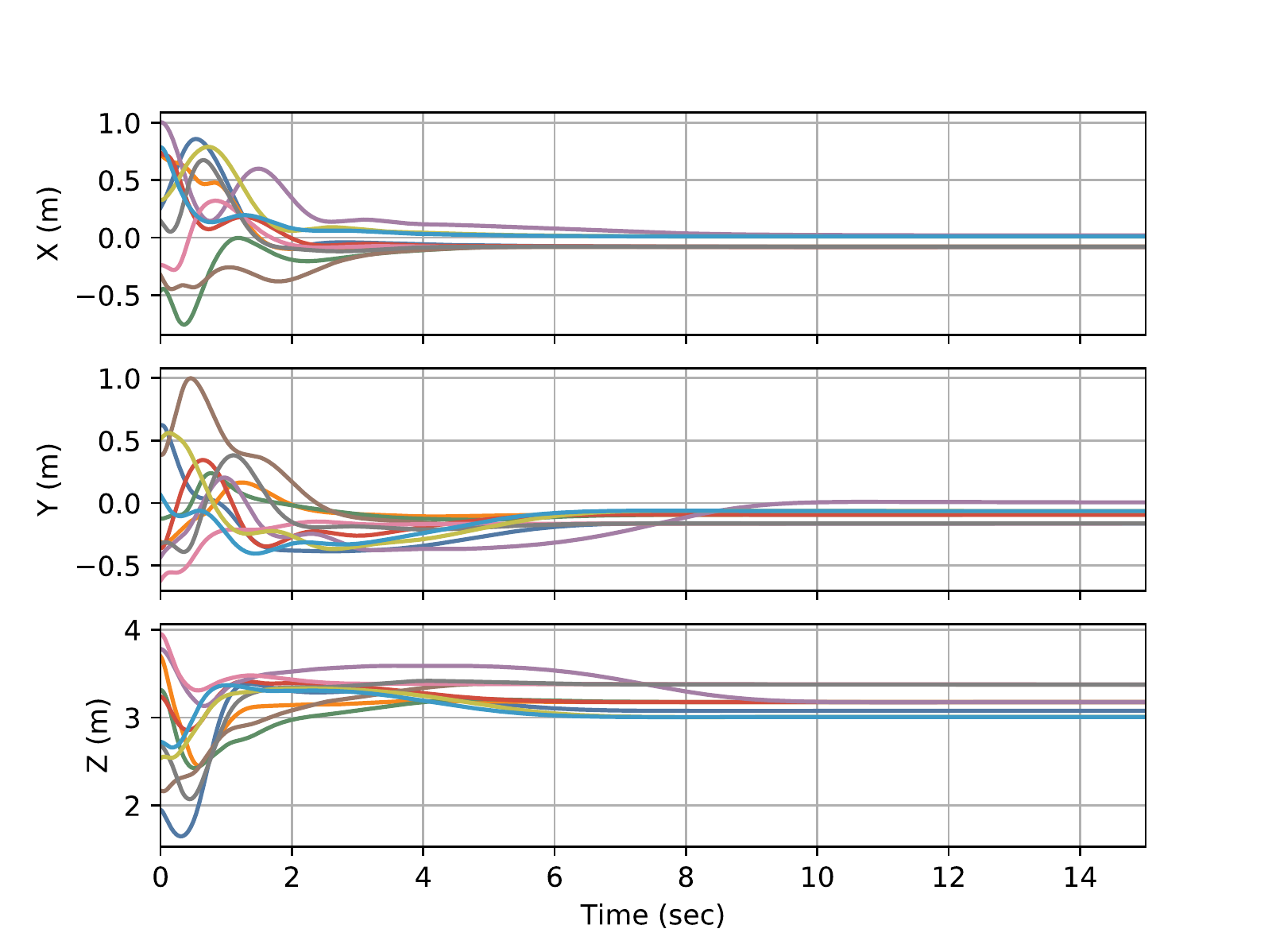}
}
\subfloat[Variation of Euler angles \label{euler-plot}]{
  \includegraphics[width=0.47\textwidth, height=0.35\textwidth]{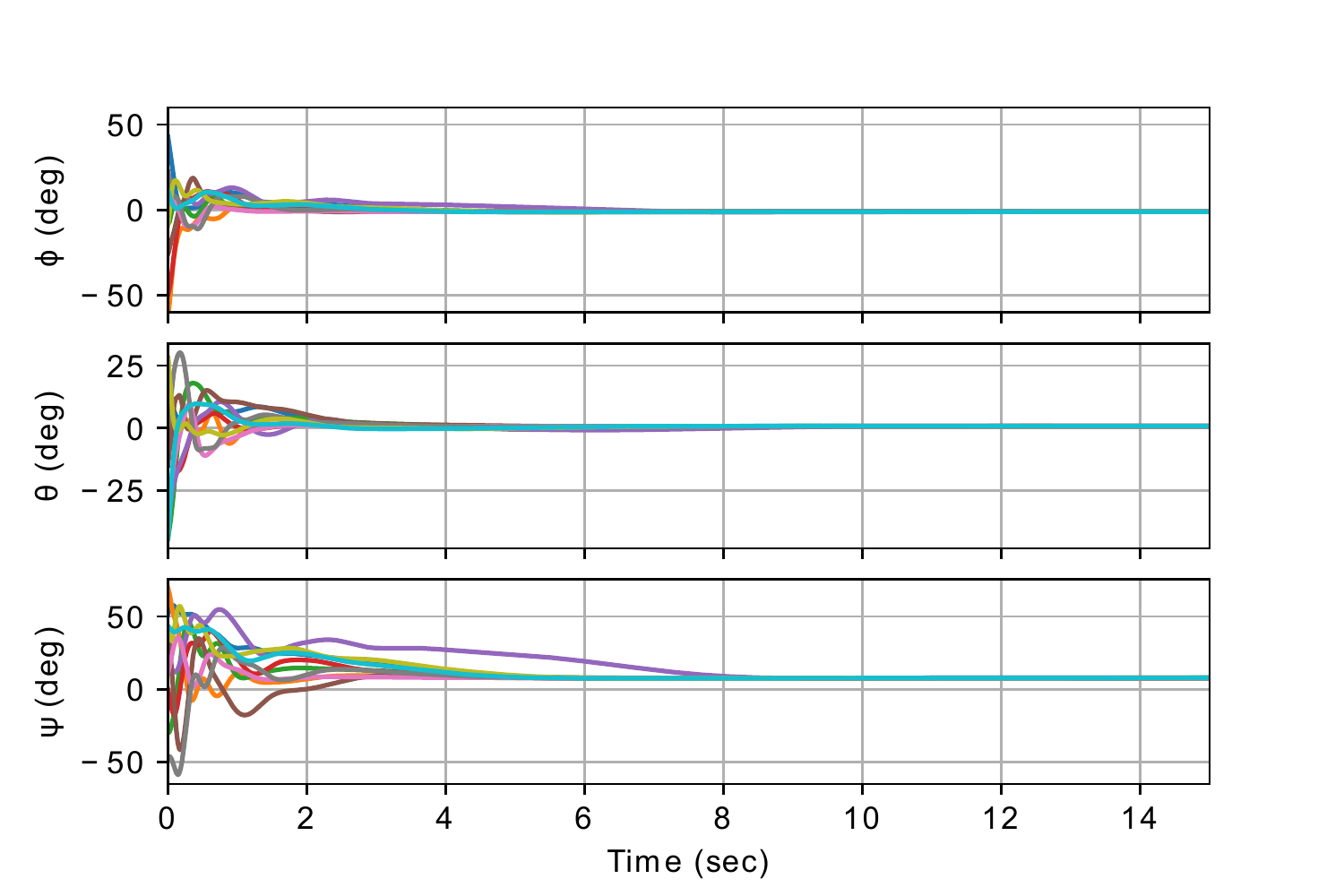}
}
\hspace{0mm}
\subfloat[Variation of tilt angles \label{tilt-plot}]{
  \includegraphics[width=0.44\textwidth, height=0.42\textwidth]{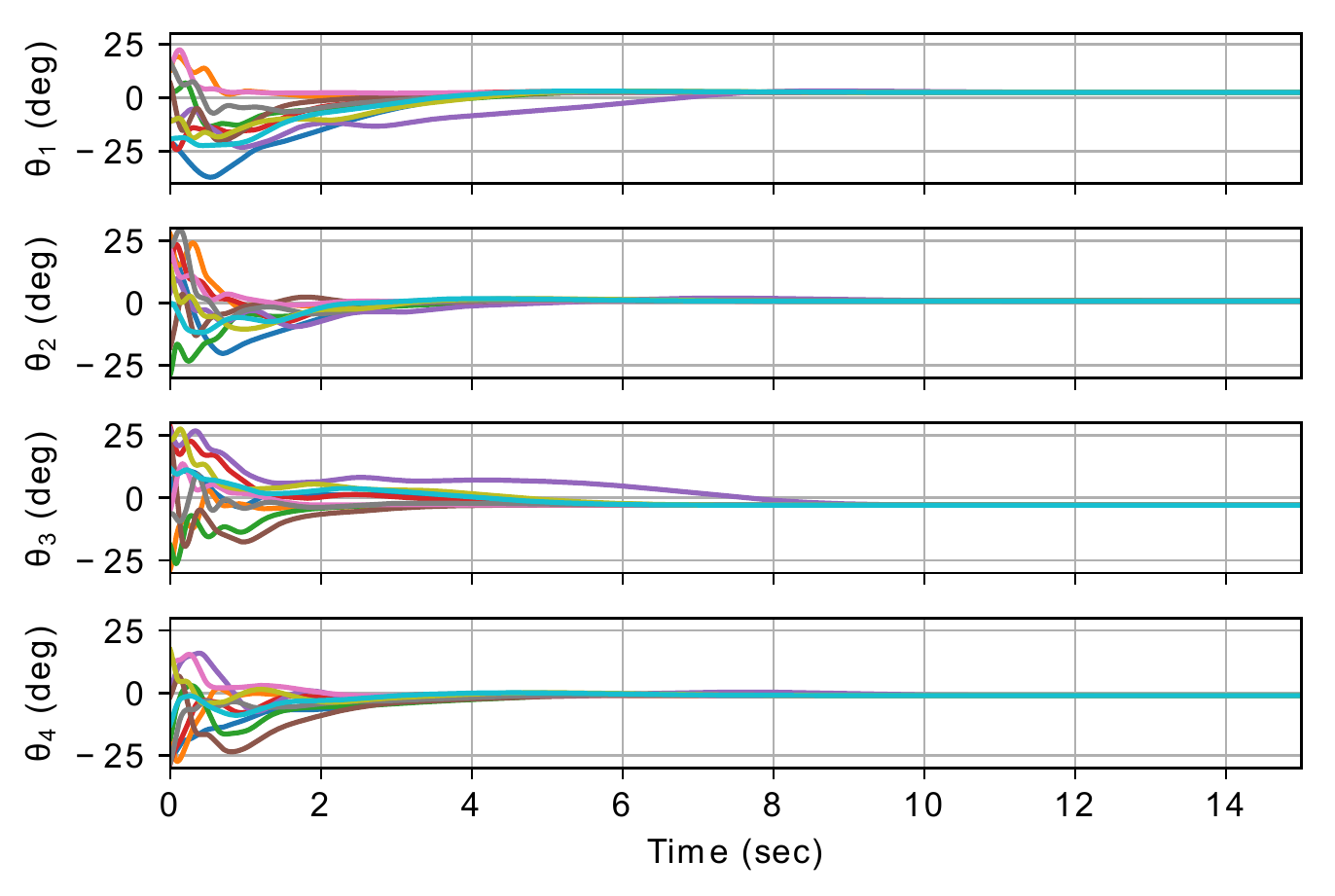}
}
\subfloat[][\centering Waypoint navigation of tilt-rotor using learned controller with developmental approach (blue) and PID-controller (orange) \label{trajectoryPID-plot}.]
{
  \includegraphics[width=0.5\textwidth, height=0.45\textwidth]{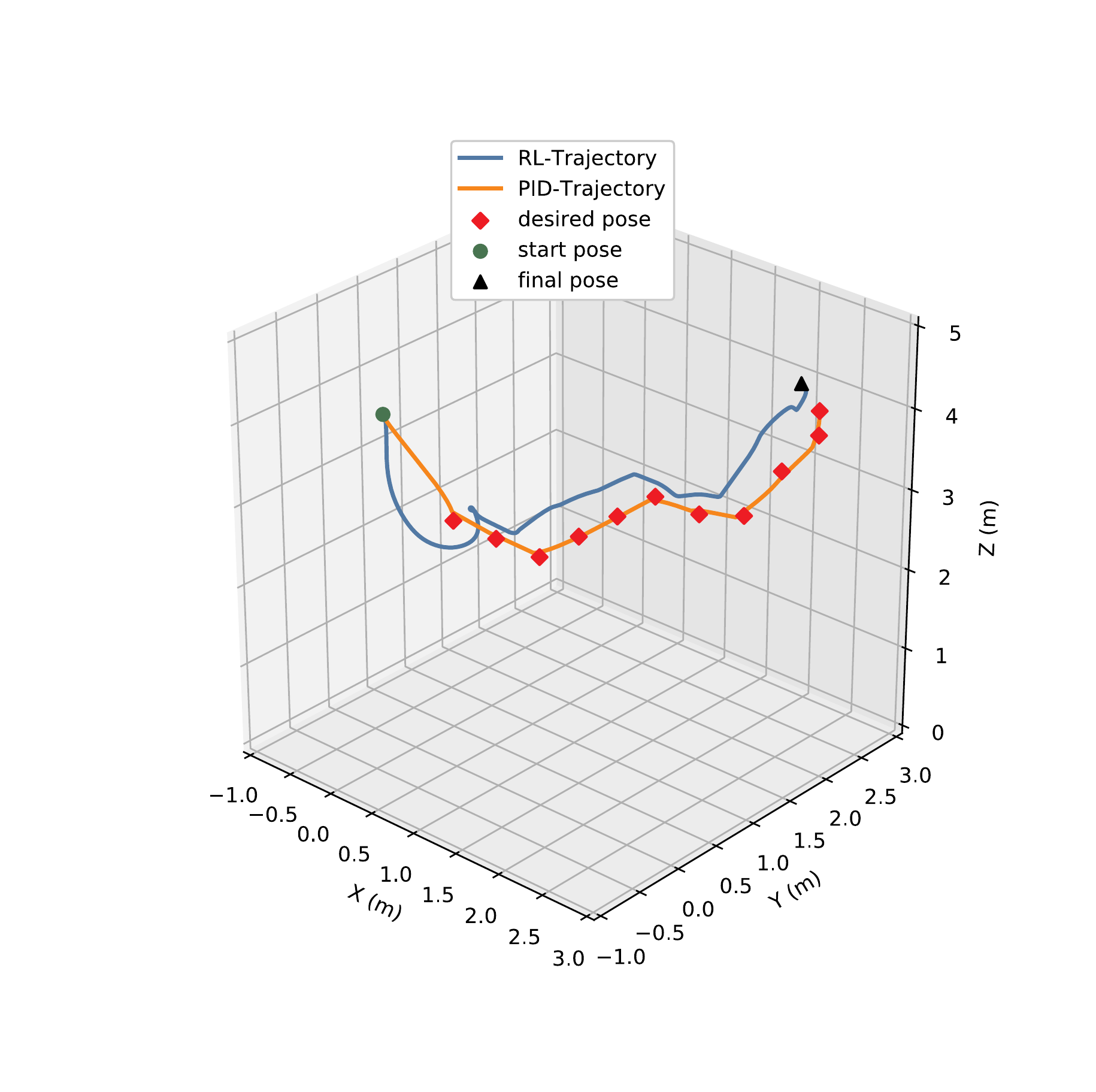}
}
\vspace{-5pt}
\caption{Results of developmental control policy tested on Tilt-rotor quadcopter.}
\vspace{-10pt}
\end{figure*}

\noindent In this section, we discuss the results obtained by the proposed approach. We compare the results of the tilt-rotor quadcopter trained from scratch with the tilt-rotor trained with the developmental transfer of policy using reinforcement learning. We refer to the results pertaining to the policy trained from scratch as $Conventional-NN$ and the results of policy trained using developmental transfer with the release of additional $dof$s on the platform in stages as $Developmental-NN$ (see section \ref{section-policy-transfer}). The neural network policy architecture used for tilt-rotor UAV in both cases was kept identical. While testing, we do not require the critic network. Only the actor network is used while deploying the control policy on the UAV. The networks used in this work are comparatively small and while running on CPU it took approximately $0.004 sec$ for each iteration in simulation.

\vspace{-12pt}

\subsection{Training Results}
\noindent Figure \ref{reward_plot}, shows a comparison of $Developmental-NN$ and $Conventional-NN$. Training for each case was performed $5$ times with different random seeds in each case for the tilt-rotor control policy and the corresponding training curves are presented in this figure.
We notice the acceleration in learning when the system is trained using developmental RL strategy.
This acceleration was achieved when the tilt-rotor control policy was trained after transfer of the learned weights from the learn quadcopter policy as described in Section \ref{section-policy-transfer}. 
This can be thought of as the policy being able to remember what it had learned in the case of the conventional quadcopter and effectively using that knowledge during learning to control the tilt-rotor platform. For a fair comparison, the number of iterations to train both the policies were kept equal. Another interesting observation which can be seen from the figure \ref{reward_plot} is that the $Developmental-NN$ was able to learn a comparatively better policy consistently during its training and it was able to achieve higher rewards as compared to $Conventional-NN$.

\vspace{-14pt}

\subsection{Ablation Study of Tilt-Rotor Actuators}
\noindent In this test, we compared the performance of the $Developmental-NN$ policy with the $Conventional-NN$ in the presence of tilt-rotor servo faults. Here, we studied four cases of failures. The results are presented in table \ref{ablation_table}, with each case is presented in a column. A faulty servo is modeled as one which responds to the commanded policy input with a probability of $0.4$. Each test was performed 100 times on both platforms. In each case, the UAV was initialized randomly using the procedure described in section \ref{section-ppo}. The initial states of UAV for each trial performed on $Developmental-NN$ and $Conventional-NN$ were kept identical for a fair comparison. The servos which were disabled were also kept identical in each trial for both the cases. The servos to be disabled were chosen randomly for each trial. The policy was considered successful if it was able to reach the desired waypoint within 1500 timesteps. The desired waypoint considered in this case was $(0, 0, 3.0)$ and the UAV was initialized within $1m$ distance of this location. The waypoint was considered as reached if the position of the vehicle was within $0.2m$ of that point.
\newline
\indent The results in the table \ref{ablation_table} as well as the fig. \ref{reward_plot} suggest that the $Developmental-NN$ policy is superior in performance as compared to $Conventional-NN$. Developmental learning allowed the trained RL agent to effectively bootstrap the knowledge it gained from the simpler design of the platform and thus it was able to complete the task more easily. It effectively exploited the learned ability of the policy trained for quadcopter (fewer $dof$s) to fly. This is also a likely reason that the learned policy is tolerant to faults in tilt-rotor servos. The difference in performance observed in column $4$ of the table \ref{ablation_table} is small. As all the four tilt-rotor servos have failed in this case, both the policies were not able to control tilting-servos effectively.

\begin{table}[ht!]
    \centering
    \vspace{-0pt}
    \setlength{\extrarowheight}{2pt}
    \caption{Ablation study: successes in 100 trials using tilt-rotor \label{ablation_table}}
    \vspace{-7pt}
    \begin{tabular}{||p{0.35\linewidth}||p{0.07\linewidth}|p{0.07\linewidth}|p{0.07\linewidth}|p{0.07\linewidth}||}
         \hline
         \multirow{3}{*}{Policy} & \multicolumn{4}{c||}{\# of faulty servos in tilt-rotor} \\\cline{2-5}
                                 & \multicolumn{1}{p{0.07\linewidth}|}{$1$} & \multicolumn{1}{p{0.07\linewidth}|}{$2$} & \multicolumn{1}{p{0.07\linewidth}|}{$3$} & \multicolumn{1}{p{0.07\linewidth}||}{$4$} \\ 
         \hline \hline
         \ttfamily $Conventional-NN$ & \ttfamily $80$ & \ttfamily $66$ & \ttfamily$31$ & \ttfamily $19$ \\ [-1.0ex]
         \ttfamily $Developmental-NN$ & \ttfamily $\mathbf{92}$ & \ttfamily $\mathbf{82}$ & \ttfamily $\mathbf{49}$ & \ttfamily $\mathbf{24}$ \\
         \hline
    \end{tabular}
    \vspace{-20pt}
\end{table}

\vspace{-10pt}

\subsection{Flight Simulation Results}
\noindent We evaluated the performance of the $Developmental-NN$-based control policy of the tilt-rotor UAV for two cases. The first case considers random initialization of the UAV in three dimensional space and the UAV is required to reach a desired goal location. In the second case, the performance of the RL-based controller is compared against a PID-based controller. The goal position is set to [0, 0, 3$m$] for the first case. Figure \ref{trajectory-plot} shows the $3D$-trajectory plots of 10 different trials. It should be noted that the UAV is initialized at random arbitrary states and it always converges towards the desired goal location as shown by the final pose cluster in figure \ref{trajectory-plot}. The corresponding time series variation of the UAV pose is illustrated in fig. \ref{xyz-trajectory-plot}. The UAV initializes at random arbitrary locations, yet it is always able to reach the vicinity of the desired goal using the $Developmental-NN$ policy control. Figure \ref{euler-plot} shows the variation of Euler angles during flight simulations. The UAV initializes at very large orientation angles as if thrown by the user. The orientation angles $\phi, \theta, \psi$ are successfully controlled by the RL-based controller. The Euler angles are well damped and approach $0^{o}$ at the hover condition when the UAV reaches the desired goal. Figure \ref{tilt-plot} shows the variation of rotor-tilt angles. The RL-based controller achieves efficient control allocation and coordination in thrust vectoring of different rotors as the system is driven towards the goal location. The rotor-tilt angles minimize to $0^{o}$ at the hover condition of the quadcopter.
\newline
\indent The performance of the RL-based controller was found to be very comparable to a conventional PID-based controller for the tilt-rotor UAV. In second case, the UAV is commanded to perform waypoint navigation across a set of predefined waypoints using $Developmental-NN$-based flight control. Figure \ref{trajectoryPID-plot} shows the three dimensional trajectory plot for waypoint navigation. The UAV is initialized at a random arbitrary state and it has has to perform a waypoint navigation mission. The PID-based control for the tilt-rotor UAV was implemented as described in \cite{kumar2018reconfigurable}. It can be observed that the tilt-rotor UAV with only RL-based controller successfully visits all the waypoints.

\vspace{-15pt}

\section{Conclusion and Future Work}
\noindent This paper introduced developmental reinforcement learning to facilitate the control of a tilt-rotor UAV with a high-dimensional continuous action space. The results indicate that the presented learning approach allowed faster and more effective learning of control in the tilt-rotor UAV platform than the conventional learning approach. The ablation studies demonstrated the fault tolerant behavior of the learned policy using a developmental approach. This result suggests that the trained policy is robust to faults. The presented work demonstrates the basic premise of developmental learning that learning carried out by gradual addition of action space and observation space can be more effective and faster than training the full system from scratch.

To allow the transition of the learned policies from simulation to real world, this work can be extended to include domain randomization \cite{sim2multireal2019}. This will allow the trained policies to be tested and deployed on real hardware platforms. Future work can also include transfer of policies to different UAV designs with more involved morphological variations.

\vspace{-17pt}

\bibliographystyle{IEEEtran}
\bibliography{bibliography}
\end{document}